# Dynamic Multi Objective Particle Swarm Optimization based on a New Environment Change Detection Strategy

Ahlem Aboud, Raja Fdhila and Adel M. Alimi

REGIM-Lab.: REsearch Groups on Intelligent Machines, University of Sfax, National Engineering School of Sfax (ENIS), BP 1173, Sfax, 3038, Tunisia
`{aboud.ahlem.tn, raja.fdhila, adel.alimi}@ieee.org`

**Abstract.** The dynamic of real-world optimization problems raises new challenges to the traditional particle swarm optimization (PSO). Responding to these challenges, the dynamic optimization has received considerable attention over the past decade. This paper introduces a new dynamic multi-objective optimization based particle swarm optimization (Dynamic-MOPSO).The main idea of this paper is to solve such dynamic problem based on a new environment change detection strategy using the advantage of the particle swarm optimization. In this way, our approach has been developed not just to obtain the optimal solution, but also to have a capability to detect the environment changes. Thereby, Dynamic-MOPSO ensures the balance between the exploration and the exploitation in dynamic research space. Our approach is tested through the most popularized dynamic benchmark's functions to evaluate its performance as a good method.

**Keywords:** dynamic optimization, dynamic multi-objective problems, particle swarms optimization, dynamic environment, time varying parameters.

## 1 Introduction

In previous years optimization problems are limited below static research space related to various applications [1-7] and many other works are based on meta-heuristics such as the swarm intelligence approach, including particle swarm optimization (PSO), ant colony optimization (ACO) and bee-inspired methods [8-16] .The domain of optimization presents a big challenge to resolve single or multi-objective problems in order to maximize or minimize the fitness function. The field related to evolutionary multi objective optimization has a several amounts of research interest in many real-world applications, to resolve relatively two objectives or more that has in conflict with one another. Due to multi-objectivity, the goal of solving Multi-Objective Problems (MOPs) isn't always discovering one optimal solution but a set of solutions. Although dynamic and multi-objective optimization have separately received an immense interest. In the literature, dynamism tasks are related to the objective function, constraints and the parameters of a predefined problem that change over the time. Dynamic multi-objective problems pose big challenges associated with the evolutionary computation approaches [17-20]. Hence, the PSO methods are proven as a good technique to solve a single and multi-objective problem in a static environment. Adapting MOPSO to resolve such as the problem is not yet treated; this is why our paper is presented.

The remaining work is delineated as follows: Section 2 presents an overview of the elementary concepts of the dynamic optimization domain. Section 3 describes the trends of dynamic optimization methods followed by our suggested approach called the Dynamic-MOPSO in Section 4. A comparative study is provided to be the topic by Section 5 followed by discussion part. The paper outline with a conclusion and some suggested ideas for future work in Section 6.

## 2      Overview of Dynamic Multi Objective Optimization Problem

The process of dynamic multi-objective optimization problems (DMOOP) [21, 22] is totally different from the static MOOP [12, 13]. In most cases, we notice that the new definition of optimality needs is to adjust a set of optimal solutions at each instance of time. A dynamic optimization problem can be defined as a dynamic problem $f_t$ such as presented by the mathematical presentation (see Equation 1: example of the minimization problem), which needs an optimization approach D, at a given optimization period $[t^{begin}, t^{end}]$, $f_t$ is termed a dynamic optimization problem in the research period $[t^{begin}, t^{end}]$ if during this period the underlying objective landscape that D helps to represent $f_t$ changes and D has to reply to this change by providing new optimal solutions.

$$\text{Min } F(x, k(t)), x = (x_1 \ldots x_n), k(t) = (k_1(t), \ldots, k_{nm}(t)) \qquad (1)$$
$$\text{Where: } g_i(x, t) \leq 0, i = 1 \ldots n_g \quad \text{and} \quad h_j(x, t) \leq 0, j = n_g + 1 \ldots n_h$$
$$x \in [x_{min}, x_{max}]$$

The main goal of the dynamic optimization approach can be explained as the problem of locating a vector of decision variables $x*(i, t)$, that is presented to be a Pareto optimal solution based on the Pareto dominance relation between solutions (see Equation 2), that satisfies an absolute set called the Pareto optimal set of solutions at instance, t, denoted as POS (t)* (see Equation 3) and improve a function vector whose dynamic values represent the best solutions that change over the time.

$$f(j, t) \prec f(i, t) * \setminus f(j, t) \in F^M \qquad (2)$$
$$POS(t)* = \left\{ x_i^* | \nexists f(j, t) \prec f(\chi_i^*, t) *, f(\chi_j, t) \in F^M \right\} \qquad (3)$$

The generated Pareto Optimal Front at time t, denoted as $POF(t)*$ is the set of the best solutions with respect to the objective space at time step t, so when solving DMOOPs the purpose is to detect the change of the best optimal front at every time instance such as defined in Equation 4:

$$POF(t)* = \left\{ f(t) = f_1(\chi^*, t), f_2(\chi^*, t), \ldots, f_{n_m}(\chi^*, t) \right\}, \forall \chi^* \in POF^*(t) \qquad (4)$$

## 3      Trends of Dynamic Optimization Approaches

In previous works, Farina *et al* [18] suggested four types to categorize the DMOOPs which outlined in table (see Table 1). Many works are done, including various types and one of the active research areas in the last few years is Evolutionary Dynamic Optimization (EDO) domain [23, 24]. Many researchers have been highlighted the intention in evolutionary computation (EC).

**Table 1.** Categories of DMOOPs.

| | Pareto Optimal Set (POS) | |
|---|---|---|
| **Pareto Optimal Front (POF)** | **No Change** | **Change** |
| **No Change** | Type IV | Type I |
| **Change** | Type III | Type II |

Many real-world problems are time-dependent parameters that involve optimization in a dynamic environment [25, 26]. To deal with the various changes in the environment, many EDO methods take into consideration as a reactive strategy. In this context, we are facing two conditions: either the algorithm has to define a methodology to find the change in the environment, or made it well known before the optimization process. To detect changes in the environment, we have typically followed one of the particular consecutive approaches: the first one is detecting modification changes by re-evaluated detectors or, the second, detecting changes based on a set of the state behaviors defined by the algorithm itself.

Many methods are related to the first strategy that introduces the change detection as a process of re-evaluating frequent existing solutions. In order to follow the previous context, function values and their feasibility must be a part of the swarm. Some existing optimization methods manage separately the detectors in the search population to ensure flexibility to maintain a high convergence during the run time and to ensure the exploitation in the research area. As a result, many diversity-based approaches is implemented such as the Dynamic Non-dominated Sorting Genetic Algorithm II (D-NSGA-II) [27], the Dynamic Constrained NSGA-II (DC-NSGA-II) [28] the individual Diversity Multi-objective Optimization EA (IDMOEA) and others [29]. One of re-evaluating detectors advantage is ensuring robustness in the time-varying environment. To maximize the performance of the algorithm, an important number of detectors are used to entail additional function evaluations. As a consequence, the used methodology requires to become informed of the most optimal number of used detectors.

To detect changes based on the behaviors of the algorithm, researchers must define a monitoring method to calculate the average of the best optimal solution founded over the time. The benefits of this method, are there are no detectors and does not require any additional function evaluations. Because no detector is used, but there may be no support that assures changes are detected and the algorithm response unnecessarily when no change occurs [30]. Many others approach is treated to predict change parameters in the environment [31].

## 4     The Proposed Approach of the Dynamic-MOPSO Based on a New Environment Change Detection Strategy

The Dynamic Multi-Objective Particle Swarm Optimization denoted by the Dynamic-MOPSO is developed based on the advantage of the fashionable particle swarm optimization technique that was in 1995 developed by Kennedy and Eberhart [12], when every particle in the population represents a candidate solution and characterized by specific parameters to be optimized in the quest research process by way of updating their position (see Equation 5) and velocity (see Equation 6) at each generation. PSO is an ideal evolutionary computation approach which can be able to resolve single and multi-objective problems for static search spaces.

$$X(k+1) = X(k) + V(k+1) \qquad (5)$$

$$V(k+1) = w * v(k) + c_1 * \text{rand}() * (p_{id}(k) - X(k)) + c_2 * \text{rand}() * (p_{gd}(k) - X(k)) \qquad (6)$$

As far as, the use of the standard MOPSO as an optimization method for dynamic problem has many negative consequences for the constraints, number of variables or their domain and the objective function of the defined problem and it can cause the problem of stagnation in local optima and many solutions can disappear over the time and cause the problem of lack of diversity and convergence after each change. As consequences, the MOPSO cannot be carried through to dynamic environment without any modifications to keep swarm diversity. In order to resolve a dynamic problems, a specific approach will be able to identify when a change in the dynamic research space has taken place after which react to such change to track the most beneficial set of solutions and to adapt in the new modified environment and this why our approach is developed. Our motivation is presented through the architecture in figure (see Figure 1) that is investigated in type I of DMOOPs, when the problem has a change in the optimal decision variables $x_i(t)^*$ when the optimal objective function does not change.

Two major problems should be resolved to handle the changed parameters: first, the way to discover that a change has occurred, second the way to respond or react appropriately to the change. Our proposed approach started with the process of MOPSO, considering the example of Dynamic Multi-Objective Problem (DMOP) such as presented in the above mathematical presentation (see Equation 1). Our proposed technique is developed to resolve a dynamic multi objective problem with dynamic parameters. In this paper, we specifically consider the environmental change that may have an effect in the parameters of the problem after each interval of time the problem which is defined as follows: $t = \frac{1}{n_t} \left\lfloor \frac{\tau}{\tau_t} \right\rfloor$; where $n_t$, $\tau_t$ and $\tau$ represent the severity, the frequency of change, and the iteration counter, respectively.

The Dynamic-MOPSO presents two main steps which are the dynamic detection and the reaction strategy that will be detailed step by step, then after the evaluation step of the fitness function $F(i)$ of each particle $p$ in the swarm $S$, our proposed approach presents an environment change detection strategy that can be able to detect change based on the process to re-evaluate the set optimal solutions $POF(t)^*$ that evaluates each candidate solutions $x$ after each interval of time $\tau_t$.

**Environmental change detection strategy:** is the first step that aims to detect the change in the research space and we preserve the set of non-dominated solutions denoted by $POF(t)^*$ as a detector of the observed change that is caused by the influence of the parameters change. As a result, the detection of dynamic parameters after each interval of time presented by the iteration counter, in our case the interval of time is defined to $(t=10)$ this parameter presents the speed of change which is severe when the value is small, moderate if frequency and severity are equal, and slight environmental changes when the value is very high. In our case, the frequency of change is defined as the value of severity to ensure a moderate environmental change.

**The response change strategy:** are the second and the important step that is elaborated to maintain convergence and diversity in the research area. After change detection steps, a tested process is implemented to verify the number of false negative changes of optimal solutions , after each generation cycle, we need to compute the number of individuals in the population which has a negative change in the value of the fitness function F(i) . In our algorithm the reactive strategy is defined by the re-initialization of all the solutions (particles) that presents a negative change in the non-dominated solution, to ensure that this dominated solution cannot lead particles to trap in local optima. Another primordial step is to re-evaluate the archive in order to update the best optimal solution at each time instance until the end of the optimization process.

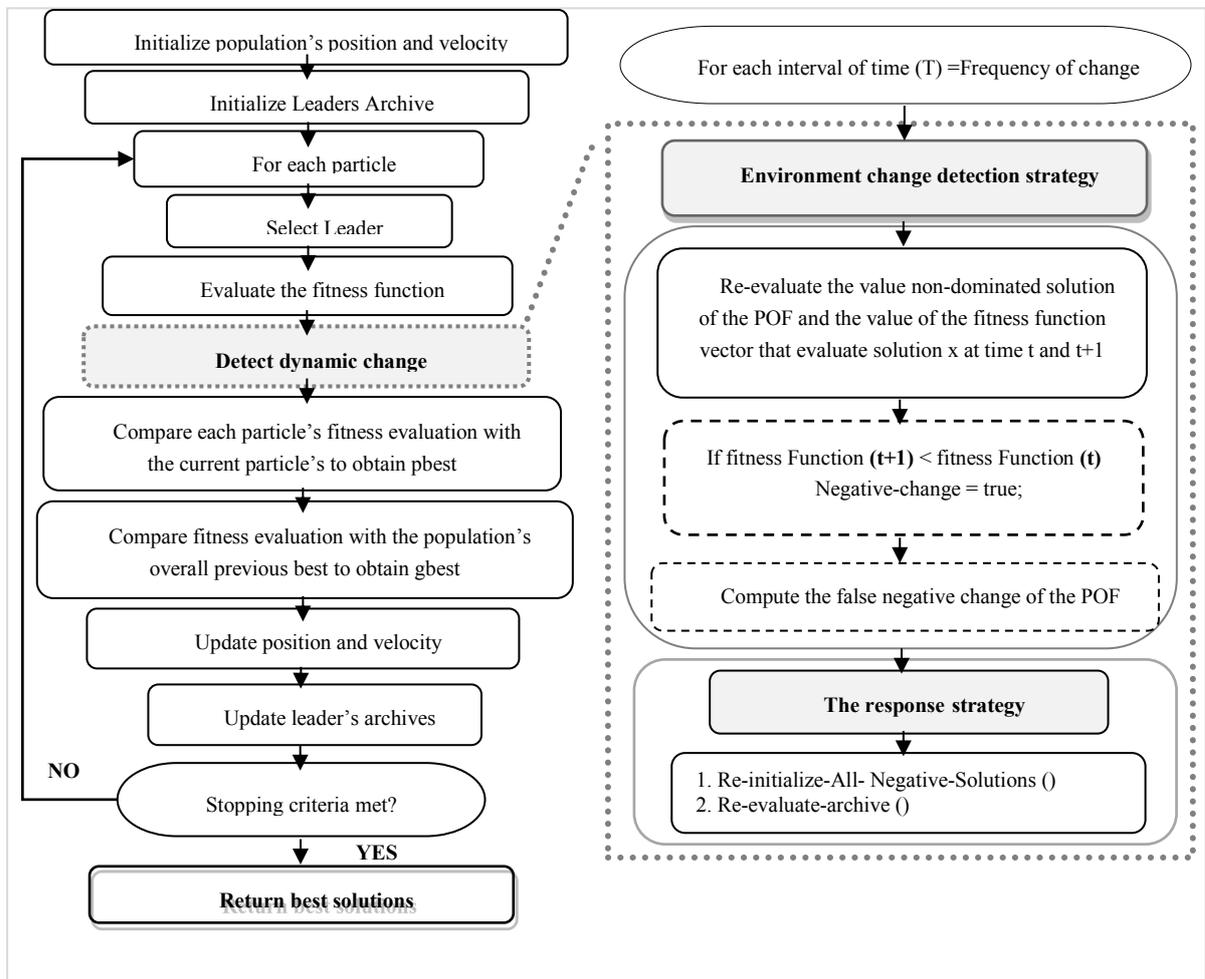

**Fig. 1.** Proposed Approach of Dynamic Multi Objective Particle Swarm Optimization

## 5     Experiments and Results

Our experimental studies based on the benchmark functions of the FDA, DIMP and dMOP suites functions, which are classified into the type I of dynamic multi-objective optimization problem [18].

### 5.1     Parameters Setting

The parameterization used for testing algorithms details in table 2:

**Table 2.** Parameters Setting

| Parameters | |
| --- | --- |
| **Common parameters of dynamic-MOPSO and OMOPSO** | Swarm size=200<br>Archive size=100<br>Independent runs =30<br>Mutation probability= 1.0/ number of problems' variable<br>Acceleration Coefficients (c1, c2) = Rand (1.5, 2.0)<br>Inertia weight (w) = Rand(0.1,0.5)<br>Max iteration = 200 |
| **Parameters of NSGAII** | Swarm size   = 100<br>Max iteration = 25000<br>Crossover Probability    = 0.9<br>Mutation Distribution Index = 20<br>Crossover Distribution Index = 20 |

### 5.2     Performance Metric

Each dynamic optimization process needs to be evaluated adopting a quality indicator to maintain diversity using the spread ($\Delta$) as a performance metrics, the generational distance (GD) to measure the convergence and the hyper-volume (HV) to measure both of them.

- **The GD:** used to measure the convergence of the approximated best solutions towards the true POF. The GD is defined in the equation 7:

$$GD = \frac{\sqrt{\sum_{i=1}^{n_{POF*}} d_i^2}}{n_{POF*}} \tag{7}$$

- **The Spread ($\Delta$):** The metric $\Delta$ measures the diversity between consecutive solutions inside the Pareto front PF. Mathematically, $\Delta$ is presented in equation 8:

$$\Delta = \sum_{i=1}^{|POF|} \frac{|dist_i - \overline{dist}|}{|POF|} \tag{8}$$

- **The HV or S-metric:** computes the scale of the location that is dominated by a set of non-dominated solutions, based on a reference vector. Mathematically, HV is defined in equation 9:

$$HV = U_i \, vol_i \, | i \in POF \tag{9}$$

### 5.3    Results and Discussion

The present section is yielding to analyze the results of the experiments and to evaluate the effect of environment change parameters defined by means of the severity (nt) and the frequency (τt) of change was both set to 10.We started out by producing the qualitative results, such as presented in figure (see Figure 2) which gives the shape of the Pareto optimal fronts and shows that adapting MOPSO to dynamic environment without take in consideration of the parameters change cause a problem that all solutions converge to the local optima. So, the dynamic-MOPSO cover this problem, the figure shows that our method are more effectively and correctly to converge to the true POF and keep diversity in dynamic  research space.

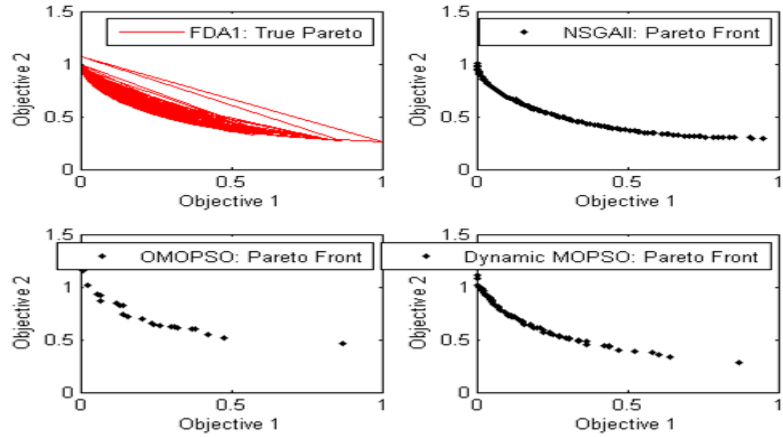

**Fig. 2.** The Generated Pareto fronts of NSGAII, OMOPSO and Dynamic-MOPSO Algorithms for FDA1 Function.

To more explain the performance measure of our approach, the quantitative results are generated applying quality indicator such as the GD, the Spread and the HV respectively. From the results shows in table 3, we can be concluded that our approach which called the Dynamic-MOPSO obtains the leading results that are highlighted in bold face for the tested FDA1 and dMOP3 problems compared with the standard OMOPSO and the NSGAII algorithms.

**Table 3.** Quantitative Results of Tested Approach OMOPSO, NSGAII and Dynamic-MOPSO

| DMOOPs | Quality Indicators | OMOPSO | NSGAII | Dynamic-MOPSO |
|---|---|---|---|---|
| **FDA1** | GD | 2.68e | 2.29e | **1.32e** |
| | Δ | 7.21e | 7.27e | **3.94e** |
| | HV | 5.57e | 5.81e | **7.74e** |
| **DIMP2** | GD | 4.14e | **1.19e** | 3.13e |
| | Δ | 1.76e | 1.60e | **1.19e** |
| | HV | **5.36e** | 1.51e | 3.55e |
| **dMOP3** | GD | 4.69e | 2.95e | **1.24e** |
| | Δ | 7.68e | 8.83e | **5.92e** |
| | HV | 3.57e | 1.87e | **5.24e** |

The following figures (see Figure 3 and Figure 4) present the performance of HV over FDA1 and dMOP3 functions respectively. These two figures show that our approach can achieve a good trade-off between convergence and diversity for solving dynamic multi-objective problems.

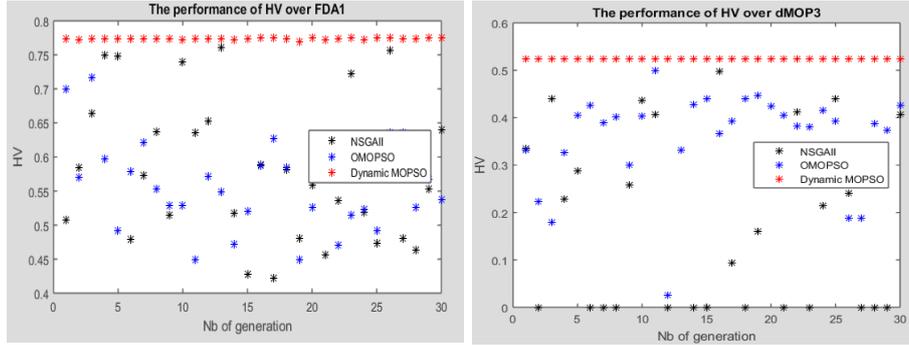

**Fig. 3.** The performance of HV over FDA1.   **Fig. 4.**  The performance of HV over dMOP3.

At the conclusion of the experimental studies, the OMOPSO and NSGAII algorithms are an important method which might be used in previous works, however, cannot be applied to dynamic environments without any changes for that reason the Dynamic-MOPSO system is presented as a new approach to deal with this problem and the previous results proved that our method can keep exploration and exploitation during the optimization process in a dynamic environment.

## 6    Conclusions and Future Research Directions

As a conclusion, the Dynamic-MOPSO approach is implemented to solve the problem categorized into the first type of dynamic multi-objective problems. So, the traditional OMOPSO is a simple and easy algorithm, but cannot have the ability to resolve DMOOP without any modification to keep swarm exploration and exploitation in a time varying environment. The no-adaptation of MOPSO for dynamic change can cause lack of convergence and diversity in the research space. But the distinction is that the new Dynamic-MOPSO approach overcomes this limitation and achieve the goal of high convergence precision and diversity by combining the simplicity of MOPSO and their efficiency to optimize dynamic problems. As a future work, we have to enhance the proposed approach that will be used as an efficient global search technique to address feature selection tasks to optimize the online learning process.

**Acknowledgements.** The research leading to these results has received funding from the Ministry of Higher Education and Scientific Research of Tunisia under the grant agreement number LR11ES48.